# Application of Word2vec in Phoneme Recognition

Xin Feng
Beijing University of
Posts and Telecommunications
Haidian District
Beijing, China
086-15203939099
fengxin_bupt@163.com

Lei Wang
Beijing University of
Posts and Telecommunications
Haidian District
Beijing, China
086-010-62282720
wanglei_elf@bupt.edu.cn

## ABSTRACT
In this paper, we present how to hybridize a Word2vec model and an attention-based end-to-end speech recognition model. We build a phoneme recognition system based on Listen, Attend and Spell model. And the phoneme recognition model uses a word2vec model to initialize the embedding matrix for the improvement of the performance, which can increase the distance among the phoneme embedding vectors. At the same time, in order to solve the problem of overfitting in the 61 phoneme recognition model on TIMIT dataset, we propose a new training method. A 61-39 phoneme mapping comparison table is used to inverse map the phonemes of the dataset to generate more 61 phoneme training data. At the end of training, replace the dataset with a standard dataset for corrective training. Our model can achieve the best result under the TIMIT dataset which is 16.5% PER (Phoneme Error Rate).

## 1. INTRODUCTION
End-to-end speech recognition is one of the recent rapid developpments in pattern recognition, which translates the speech to text without alignment between the speech frames and characters. The traditional hybrid systems (GMM-HMM, DNN-HMM) are trained separately in several parts of the system (i.e. acoustic model; language model)[1,2,3,4]. Unlike these traditional hybrid systems, end-to-end speech recognition model simplifies the entire framework, which puts the entire model into one network structure. There are two main end-to-end speech recognition architectures: Attention-based speech recognition model aligns the speech and the recognized characters through attention mechanism [5,6,7]. CTC (Connectionist Temporal Classification) system applies Markov assumptions to solve sequence problems through dynamic processing [8,9,10]. CTC needs several independent assumptions to determine the probability of the sequence, while the attention-based model does not need to make these assumptions for decoding all the sequence. Attention mechanism was firstly introduced to speech recognition in 2015 [5]. This model is built to perform phoneme recognition experiments on different phoneme sets with the TIMIT dataset.

At the same time, a location-aware strategy is added in this basic system, which can reduce the calculation complexity of the system and improve the system performance to a certain extent. In 2016, Listen Attend and Spell system applied attention mechanism in large vocabulary speech recognition [7]. By adding a pyramid structure to the listener, the system reduces the length of the encoding sequence in listener to 1/8 of the original, which can greatly reduce the computational complexity.

In 2017, in order to solve the problem of non-monotonic alignment in the speech recognition system based on the attention model, a combination of CTC and the attention system was proposed to form a hybrid system [11,12,14]. Let CTC's monotonic alignment solve the non-monotonic problem, but this will increase the complexity of the model. After 2018, the encoder of the attention model begins to diversify. The encoder in the speech recognition system based on the attention model becomes more complicated, and CNN begins to appear in the encoder [13,14].

In the attention models, each step of decoding, the prediction result of the previous step is required. In the current decoding process, the prediction result at the previous moment is mapped to a vector, which is generated through an embedding matrix. The decoder uses this vector to predict the character together with the context vector generated by the current step decoding. In the usual model architecture, embedding matrix is often trained along with the speech recognition task. Each row in the embedding matrix represents a character / phoneme, so the distance between the embedding vectors of each phoneme / character may affect the performance of the model.

Our model is a phoneme recognition model based on attention. In the training phase, in order to fully increase the distance among the vectors in the embedding matrix, we choose to apply the phoneme's transcripts to train the embedding matrix in the word2vec system [15,17]. Then the system loads the trained embedding matrix weights to initialize the embedding matrix in the attention model. During testing, it is found that the performance of the system can be significantly improved when the vector distance is larger. We evaluate the system on a standard dataset TIMIT in our experiments [16]. TIMIT dataset is labeled based on 61 phonemes. However, for the attention system, the size of TIMIT is too small. If the model is trained using only the TIMIT dataset, it will easily get overfitting. So we propose a new method of phoneme inverse mapping for training. Through the pronunciation dictionary, randomly map the phoneme labels of the 39 phoneme dataset to 61 phonemes, and then feed them into the model for training. Finally, in the evaluation, the phonemes are mapped back to 39 phonemes for evaluation. In this paper, we use the parameters trained by word2vec to initialize the weights in attention by combining the word2vec model and the attention model. At the same time, we make a random mapping of 39 to 61 phonemes and then train the model. At the end, we apply a model corrective training.

In section 2, we detail how to do system hybridization of the two models. In section 3, we introduce how to randomly map 39 phonemes to 61 phonemes to participate in training to improve system performance. We have a detailed analysis of the experimental results, in section 4. In the end, the best evaluation result that can be obtained under the TIMIT dataset is 16.5% PER. So we demonstrated our point through experimental comparison. (Section 5)

## 2. JOINT WORD2VEC AND ATTENTION

In this section, we explain how to make the word2vec model to initialize the embedding matrix in the attention model. Section 2.3 explains in detail how the word2vec model and the attention model are hybrid.

### 2.1 Attention-based Model Encoder-Decoder

Attention model is usually an encoder-decoder architecture. Encoder encodes the audio feature sequence input $X = (x_1, x_2, ..., x_T)$ which is extracted from the speech into a high-level feature representation $\mathbf{h} = (h_1, h_2, ..., h_T)$, as follow:

$$\mathbf{h} = \text{Encoder}(X). \quad (1)$$

Each time the Decoder decodes, we can get the relativity between the hidden state $s_i$ of the current decoder and the feature code $\mathbf{h}$ by scoring:

$$e_{i,t} = score(s_i, \mathbf{h}_t), \quad (2)$$

$$\alpha_{i,t} = \frac{\exp(e_{i,t})}{\sum_t \exp(e_{i,t})}. \quad (3)$$

According to attention and feature coding $\mathbf{h}$, a context vector can be generated which is all the information needed for the current decoding step in encoder $\mathbf{h}$ [5,7], as:

$$c_i = \sum_t \alpha_{i,t} \mathbf{h}_t \quad (4)$$

$$v_{i-1} = \text{lookup}(\mathbf{M}, y_{i-1}). \quad (5)$$

The output result of the decoding at the previous moment is forwarded to the decoding at the next moment. And look up in the $\mathbf{M}$ (embedding matrix) to turn the decoding result into a vector $v$. This vector, together with the context and the hidden state $s_{i-1}$ of the decoder at the previous moment, generates the decoding result of the current step:

$$s_i = \text{Recurrency}(s_{i-1}, c_{i-1}, v_{i-1}) \quad (6)$$

$$y_i = \text{Generate}(c_i, s_i) \quad (7)$$

We can see that the embedding matrix in the decoding process is trained along the entire system. It may make it difficult to fully train the embedding matrix. So we consider taking out the embedding matrix and training it separately.

### 2.2 Embedding Matrix Trained by Word2vec

Word2vec system is a model for learning in an unsupervised manner from a large number of text corpora. The model uses word vectors to represent semantic information by learning text, which sneaks into a space to make semantically similar words very close in the space, thereby increasing the distance of different kinds of words in the space [17].

During training, the word2vec system maps the words by an embedding matrix, so that the words are mapped from the original space to the new multi-dimensional space. Two algorithms commonly used in word2vec system are: CBOW and skip-gram. The CBOW model is to input a context-dependent word vector of a certain feature word, and outputs the word vector of the intermediate word [15] . The skip-gram model is to input a word vector of a specific word and outputs a word vector of a specific word context.

In this phoneme recognition model, the model unit is phoneme. So during training, we converted all text into phoneme level through the pronunciation dictionary. Then the phoneme-level text corpus is put into the word2vec system to train the embedding matrix until the distance among all vectors is maximized.

### 2.3 Joint of Word2vec and Attention

The joint model uses a word2vec model to train the embedding matrix. Apply the parameters of this model training to initialize the embedding matrix in the attention model, then make the system to be jointly trained.

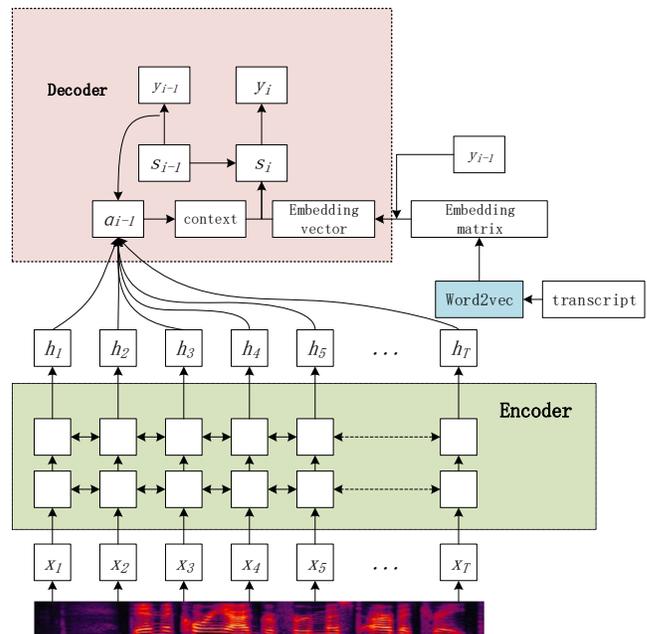

**Figure 1. Word2vec-attention joint model architecture.**

As shown in Figure 1, The entire system architecture is divided into two parts: attention model and Word2vec model. Before the attention model training, the embedding matrix has been trained in word2vec model using all the phoneme/character transcription. To improve the performance, the distance between the embedding vectors of each phoneme in the embedding matrix has reached the maximum. Our attention system is very similar to the LAS architecture [7], which is a typical encoder-decoder architecture. Figure 1 shows that at each decoding step, a vector is generated after the previous decoding result is mapped by the embedding matrix, and the embedding matrix is trained using word2vec with all transcription.

## 3. TRAINING METHOD

The standard phoneme dataset of 61 phonemes often cannot meet the need of attention model training. We design a method of inverse phoneme mapping, so that more 39-phoneme datasets are put into the 61 phoneme model for training. It can solve the overfitting problem and improve model's performance.

## 3.1 Phoneme random inverse mapping

Our system is built based on phoneme recognition. In many classic English phoneme recognition systems, there are three types of phoneme sets, 61, 48, and 39. But the standard dataset for 61-phoneme is only TIMIT. All models are trained using the 61-phoneme set, which should be mapped to 39-phoneme by the CMU 61-39 phoneme lookup table during evaluation. But there is no clear correspondence between 39 and 61 phoneme. The dictionary of the TIMIT dataset is a 61-phoneme dictionary, but the vocabulary is very small, only 7,000 words.

Howevert we have a large vocabulary 39-phoneme dictionary (CMU pronunciation dictionary)[18], which has a very large vocabulary. If we need to more 61-phoneme to transcribe audio data for training the model, we have to consider mapping the 39 phoneme labels to 61-phoneme according to the conversion dictionary. So we propose a new training method, a random inverse mapping of 39-phoneme to 61-phoneme.

**Table1: Mapping and inverse mapping 61-39 phonemes(some examples)**

| 61 to 39 | | 39 to 61 | |
|---|---|---|---|
| aa,ao | aa | aa | aa/ao |
| ah,ax,ax-h | ah | ah | ah/ax/ax-h |
| bcl/dcl… | sil | d | dcl d |

Here we prepare the mapping relationship from 61 to 39 [19]. In the face of many-to-one mapping, the phoneme can randomly extract a replacement of the original phoneme from multiple phonemes during inverse mapping. The spatial distance of these similar phonemes is not large, because when performing a system evaluation, they can be mapped to the same phoneme. Inverse mapping also includes expanding a phoneme to two phonemes (for example, d to dcl d). In this way, we can convert all 39-phoneme labeled datasets to 61-phoneme labeled datasets. When the system is evaluated after training, the 61-phonemes are still converted to 39-phonemes according to the mapping relationship for system evaluation. Through the expansion of phoneme types, 39-phoneme is inversely mapped to 61-phoneme. When they are mapped to 39-phoneme again during the evaluation phase, the distance among phonemes will be increased, thereby improving the performance of the phoneme recognition system.

## 3.2 Model initialization of phoneme random inverse mapping

Due to the lack of a 61-phoneme dataset, when training a 61-phoneme recognition system, we propose an inverse phoneme mapping method. The 39-phoneme transcribed speech dataset is inversely mapped, so that a large dataset of 61-phoneme can be obtained. We use the large dataset to train the model until the model evaluation reaches the best.

However, our random inverse mapping is not completely accurate. There is a certain degree of randomness in the process of inverse phoneme mapping. There may even be labeling errors, so we need to use standard dataset to correct the training of the model. Even though the TIMIT dataset of 61-phoneme transcription is small, the accuracy of the annotation can be guaranteed. Therefore, we can let the large dataset that has been inversely mapped to initialize the model parameters to minimize the loss of model during training. At this time, the performance of the model can be best on the large dataset, and put the standard dataset to replace large datasets for a corrective training. Let the standard dataset correct errors made with datasets transcribed using our own phoneme reverse mapping.

## 4. EXPERIMENT

### 4.1 Data and Model

Our experiment prepared two datasets: librispeech-clean-360 [20] and TIMIT. We divide librispeech into three parts, and extract 5000 audios as the test set, 2000 audios as the validation set, and the rest for training. Because our experiment is designed for two phoneme sets, 61-phonemes and 39-phonemes, we use the CMU's 39 phoneme pronunciation dictionary to transcribe all data in the librispeech dataset into 39 phoneme transcriptions. Then use the 61-39 phoneme mapping table and the random phoneme mapping strategy to convert all the transcriptions of the dataset to 39 phonemes. All labels are extracted as corpus text for training word2vec model.

As input, we extract 40 mel-scale filterbank coefficients, with their first and second order temporal derivatives to obtain a total of 120 feature values per frame. Our model is built based on the LAS system framework [7]. The encoder is a 2-layer Bidirectional Long Short-Term Memory(BLSTM) with 512 cells in each layer and direction. And we don't add pyramid structure to the encoder. We quote the Bahdanau Attention mechanism [5]. The decoder is 2-layers LSTM with 512 cells in each layer. The embedding dimension in both the attention and word2vec systems is 32, which means that each time the output of the decoder is embedded into a 32-dim vector as the input for the next decoding. Attention model training uses scheduled sampling = 0.1 [21]. The beam search was set in decoding under all conditions.

### 4.2 Result

There are 4 kinds of distance between the vectors of the embedding matrix. Cos is the cosine distance. M-dist is the Mahalanobis distance. Cov is the mean of the covariance. P-dist is the Pearson correlation coefficient. The attention models are decoded with a beam size of 10.

As shown in table 2, there are the PER of the different experiments in the evaluation and the word vector distance in the embedding matrix. There are some differences in experimental conditions between different groups of experiments. We start with different phoneme sets and compare the experimental results of 39-phoneme sets and 61-phoneme sets to verify the system performance differences under different phoneme sets. In addition, we also demonstrate whether the embedding matrix trained in advance can improve system performance. A comparative experiment was performed to verify whether the inverse phoneme mapping plus correction training can obtain better experimental results on the standard dataset TIMIT.

#### 4.2.1 Embedding initialization

First, we can get from the 1 set of experiment and the 2 set of experiment that after initializing the attention model with the embedding matrix trained by all the transcribed texts, the word vectors can be made more divergent. The experiment was trained using the librispeech 39-phoneme set, and then evaluated simultaneously using the TIMIT and librispeech dataset. After evaluation, it is found that using word2vec to initialize the embedding matrix can improve the performance of the phoneme recognition model. At the same time, the measurement of distance shows that the distance between the embedding vectors trained by word2vec is larger. Besides, the pre-trained model can accelerate convergence during training process, greatly reducing the model

Table 2: Phoneme Error Rate (PER) on Librispeech and TIMIT in different experiments.

| Index | Phoneme | Model set | Librispeech (PER) | TIMIT (PER) | Cos | M-dist | Cov | P-dist |
|---|---|---|---|---|---|---|---|---|
| 1 | 39 | No-pretrain | 12.7 | 22.7 | 0.289 | 52.00 | 2.501 | 0.359 |
| 2 | 39 | Pretrain | 8.37 | 19.1 | 0.097 | 3.92 | 0.055 | 0.149 |
| 3 | 61 | Pretrain | 8.57 | 18.5 | 0.128 | 33.85 | 0.476 | 0.162 |
| 4 | 61 | Pretrain+all dataset | 9.41 | 18.1 | 0.142 | 44.24 | 0.631 | 0.176 |
| 5 | 61 | Pretrain+corrective training | 8.84 | **16.5** | 0.136 | 12.83 | 0.178 | 0.171 |

training time.

### 4.2.2 61-39 phoneme

We inverse phoneme mapping of librispeech data, and convert the data marked by 39 phonemes to 61 phonemes through the CMU pronunciation dictionary. 3, 4, and 5 sets of experiments are all model training performed using the phonemes from the inverse mapping dataset. During the evaluation, all predicted phonemes are mapped into a sequence of 39 phonemes according to the mapping table of 61-39 phonemes. Through comparison with the experiments 1, 2 and experiments 3, 4, 5, it was found that after the 39 phoneme transcription of the librispeech dataset was reversely mapped to 61 phonemes, the standard dataset was used for evaluation, which can greatly improve. The best results of 39-phoneme training situations is 19.1%. However, the best result of 61 phoneme training is 16.5%.

### 4.2.3 Corrective training

The librispeech dataset becomes 61 phoneme after inverse mapping. But this random mapping method may produce some labeling errors on 61 phoneme recognition. To solve this problem, we set up an experiment which trained the model to the minimum loss function under the librispeech dataset, and then input the TIMIT dataset to replace the librispeech dataset for corrective training. In this way, the best PER 16.5% can be achieved under the TIMIT dataset. In experiment 4, we mixed the librispeech dataset and the TIMIT dataset, and then put them into the training of the model together. Compared with the basic model (experiment 3), the performance evaluation of experiment 4 and experiment 5 of models can be further more improved. But in experiment 5, the performance of the system for corrective training using the TIMIT dataset improves even more. This may be mainly because the number of librispeech dataset is much larger than the TIMIT dataset, which is exactly the problem to be solved using the phoneme random inverse mapping.

## 5. CONCLUSION

We applied the word2vec model in natural language processing to end-to-end speech recognition based on attention. By using the word2vec model to train the embedding matrix, the distance between the embedding vectors is increased, thereby improving the system of the speech recognition system. At the same time, we also used the random phoneme inverse mapping method to randomly map the 39-phoneme dataset into 61-phoneme, and more phoneme types were put into the system for training. This can increase the distance among phoneme all embedding vectors. After adopting such a training method, using the standard dataset for correction training, the model can get a better performance in the standard dataset. At the same time, our experimental method also greatly reduces the training time of the attention model. In the future, we can consider using a larger word embedding model to increase the distance among the embedding vectors to improve the model effect of speech recognition. The language model can also be connected after the attention model training, which can make the model performs better.[1]

---

[1] The code of this experiment is at: https://github.com/fengxin-bupt/Application-of-Word2vec-in-Phoneme-Recognition